# A hybrid Evolutionary System for automated Artificial Neural Networks generation and simplification in biomedical applications

Enrique Fernández-Blanco[1], Daniel Rivero[1], Marcos Gestal[1], Carlos Fernández-Lozano[1], Norberto Ezquerra[2], Cristian Robert Munteanu[1], Julian Dorado[1]

[1]University of A Coruna, Fac. Informatica, Campus Elviña, 15071, A Coruña, Spain. [2]Georgia Tech, College of Computing, Atlanta, USA.

**Abstract.** Data mining and data classification over biomedical data are two of the most important research fields in computer science. Among the great diversity of technique that computer science can use for this purpose, Artificial Neural Networks (ANNs) are one of the most suited. One of the main problems in the development of this technique, ANNs, is the slow performance of the full process. Traditionally, in this development process, human experts are needed to experiment with different architectural procedures until they find the one that presents the correct results for solving a specific problem. However, recently, many different studies have emerged in which different ANN developmental techniques, more or less automated, are described, all of them having several pros and cons. In this paper, the authors have focused to develop a new technique to perform this process over biomedical data. The new technique is described in which two Evolutionary Computation (EC) techniques are mixed in order to automatically develop ANNs. These techniques are Genetic Algorithms (GAs) and Genetic Programming (GP). The work goes further, and the system described here allows the obtaining of simplified networks with a low number of neurons for resolving the problems adequately. Those already existing systems that use EC for ANN development are compared with the system proposed here. For this purpose, some of the most frequently biomedical databases have been used in order to measure the behaviour of the system and also to compare the results obtained with other ANN generation and training methods with EC tools. The authors have also used other databases that are frequently used to compare this kind of method in order to obtain a more general view of the new system's performance. The conclusions reached from these comparisons indicate that this new system produces very good results, which in the worst case are at least comparable to existing techniques and in many cases are substantially better.

Furthermore, the system has other features like variable selection. This last feature is able to discover new knowledge about the problems being solved.

## 1. INTRODUCTION

Mixing data mining and biological data is always a synonym of difficult question. Computer science has developed different tools to help in such process. One of the best known and commonly accepted tools are the Artificial Neural Networks (ANNs). This technique has been used to develop new theories , or, just to analyse the data according to previous observations .

Implementing and handling Artificial Neural Networks (ANNs) can be considered an easy task. Different learning systems were successfully applied in the resolution of a large number of problems in different fields, such as classification , clustering , regression , etc. ANNs present interesting features which make them a powerful technique for complex problem solving. Those characteristics have led many researchers to use them in a large number of different environments .

Although ANNs are a very useful tools, their use also raises a series of problems. One of the main problems that can be found is on the developmental process. This process can be divided into two parts: development of the architecture and its corresponding training and validation. Determining how many neurons an ANN will have, how many layers the structure will have and how they are interconnected is what is mainly known as development of the architecture. Training that structure is referred to how to calculate and update the weight values of the connections of the architecture.

The network architecture is dependent on the problem to be resolved, and it generally makes the design process of this architecture a result of a manual process based on experience. In other words, the expert has to experiment with various different architectures until he finds one that is capable of offering good results after the training process. Therefore, the expert has to make various experiments with different architectures and train each one of them in order to be able to determine which one will be the best. This is a slow process marked by the circumstance that determining the architecture is a manual process, although recently some ANN creation techniques have been developed in a more or less automated way.

This work intends to solve this big problem of slowness in the development of ANNs focused on biomedical data. For this purpose, a hybrid EC system is proposed in which Genetic Programming (GP) and Genetic Algorithms (GAs) are mixed in order to develop ANNs without human participation. GP is used to evolve the topology of ANNs, and GAs are used to do the training of the weights of the network. These two processes are not run separately, one after the other, but alternatively during the whole development process.

The technique described in this paper was applied to some of the most used biomedical benchmark data. Therefore, the results can be compared with others already published in other works. Also, in order to show the capabilities of the system developed here, some of the "classical" machine learning databases were used to perform the experiments and the comparison with other works.

## 2. STATE OF THE ART

This section describes a review of the most important works related to this paper, including Genetic Algorithms, Genetic Programming and development of ANNs with Evolutionary Computation tools. These techniques are particularly interesting to process biomedical data because they are quite robust to noise and uncertain data. Especially the Artificial Neural Networks are especially useful when the mathematical relation about certain data is unknown but we have samples of inputs of a system and the outputs of those systems.

*2.1 Artificial Neural Networks*

Among the whole set of tools defined by Artificial Intelligence, Artificial Neural Networks (ANNs) are one of the most used and oldest ones. ANNs were firstly defined in , where the authors develop a technique inspired in how a natural neuron process information. This work define a neuron with a structure similar to Fig 1, where the neuron computes the sum of the weighted inputs and applies a function in order to generate the output.

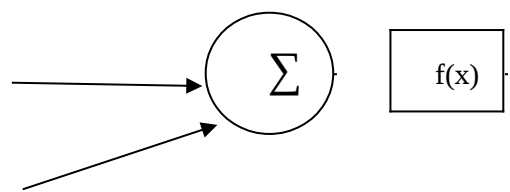

**Fig 1 Artificial Neural Network Schema**

This technique has evolved to develop more complex structures than a single artificial neuron. In literature, it can be found a large number of different architectures that have been used to solve different problems in different knowledge areas . These architectures usually contains ten or hundreds of artificial neurons which have to adjust their weights in order to collaborate and solve the problem. So, some algorithms have been developed in order to adjust the weights among artificial neurons to perform a certain task, such as, Genetic Algorithms.

*2.1. Genetic Algorithms*

Genetic Algorithms (GA) are a search technique inspired in the evolution of the species and proposed in  which are useful to solve optimization problems. From an initial random population of solutions, the population is evolved by means of three simple operators: selection, mutation and crossover. Applying those operators in an iterative process, the population reaches different states with different solutions. Those states are called generations. The result of those generations is expected to be a population that contains a solution which is good enough to solve the problem.

GAs were presented to operate over solutions codified as strings of bits, but actually they are also used with real numbers . This is an advantage, because, as it was already described in section 1 the GA will be used to train the weights of different networks. Therefore, the solutions represented in the population will codify in this form the connection weights of the network to be train.

*2.2. Genetic Programming*

Genetic programming (GP) appears as an evolution of GAs. The formal base of GP, when this technique was baptised under this name, can be dated in 1992 with the publication of the book "Genetic Programming" , although in a previous work from the same author  it was clear that genetic programming was simply an extension of genetic algorithms applied to tree-based programming structures.

In classical GP, the codification of the solutions is done with the shape of trees. For this reason, the user has to specify which terminals (leaves of the trees) and functions (nodes that have children) can be used by the evolutionary algorithms by means of a terminal and function sets. From those terminals and functions, it is possible to build complex expressions.

GP has been used on a large number of problems with great success. Although its main and most direct application is the generation of mathematical expressions , it has also been used in other different areas such as rule generation , knowledge extraction , filter design , image processing , signal analysis , etc.

*2.3. ANN development with Evolutionary Computation tools*

The development of ANNs is a topic that has been extensively dealt with very diverse techniques. The world of evolutionary algorithms is not an exception, and proof of that is the great amount of works that have been published about different techniques in this area . These techniques start from a random initial population, each individual of the population codifying different parameters (typically, the weight of the connections and / or the architecture of the network and / or the learning rules). This population is evaluated in order to determine the fitness of each individual. Afterwards, this population is repeatedly made to evolve by means of different genetic operators (replication, crossover, mutation, etc.) until a determined termination criteria is fulfilled (for example, a sufficiently good individual is obtained, or a predetermined maximum number of generations is achieved).

Essentially, the ANN generation process by means of evolutionary algorithms is divided into three main groups: evolution of the weights, architectures, and learning rules.

*2.3.1. Evolution of the weights*

The evolution of the weights begins with a network with a predetermined topology. In this case, the problem is to establish, by means of training, the values of the network connection weights. This is generally perform as a minimization problem of the network error, for example, using the Mean Square Error of the network between the desired outputs and the ones achieved by the network. Most of the training algorithms, such as the backpropagation algorithm (BP) , are based on gradient minimization. This has several drawbacks, the most important is that quite frequently the algorithm becomes stuck in a local minimum of the error function and is unable of finding the global minimum. One way of overcoming these problems is to carry out the training by means of Evolutionary Algorithm ; i.e., formulate the training process as the evolution of the connection weights in an environment defined by the network architecture and the task to be done. In these cases, the weights can be represented in the individuals' genetic material as a string of binary values or a string of real numbers . Traditional GA use a genotypic codification method with the shape of binary strings. In this way, much work has emerged that codifies the values of the weights by means of a concatenation of the binary values which represent them . The main advantage of these approximations is that it is very easy and quick to apply the genetic operators on a binary string. The disadvantage of using this type of codification is the problem of permutation. This problem was raised upon considering that the order in which the weights are taken in the string causes equivalent networks to possibly correspond with totally different individuals. This leads the crossing operator to become very inefficient. Logically, the weight value codification has also emerged in the form of real number concatenation, each one of them associated with a determined weight, which was the object of study in the 90's . By means of genetic operators designed to work with this type of codification, and given that the existing ones for bit string cannot be used here, various studies showed that this type of codification produces better results and with more efficiency and scalability than the BP algorithm.

*2.3.2. Evolution of the architectures*

The evolution of the architectures includes the generation of the topological structure; i.e., the topology and connectivity of the artificial neurons, and the transfer function of each neuron of the network. The architecture of a network has a great importance in order to successfully apply the ANNs, as the architecture has a very significant impact on the process capacity of the network. Therefore, the design of a network is crucial, and this task is classically carried out by human experts using their own experience, based on "trial and error", experimenting with a different set of architectures. The automated design of architectures has been possible thanks to the appearance of constructive and destructive algorithms . In general terms, a constructive algorithm begins with a minimum network (with a small number of layers, neurons and connections) and successively adds new layers, nodes and connections, if they are necessary, during the training. A destructive algorithm carries out the opposite operation, i.e., it begins with a maximum network and eliminates unnecessary nodes and connections during the training.

In order to develop ANN architectures by means of an evolutionary algorithm, it is necessary to decide how to codify a network inside the genotype so it can be used by the genetic operators . For this, different types of network codifications have emerged.

In the first codification method, direct codification, there is a one-to-one correspondence between the genes and the phenotypic representation . The most typical codification method consists of a matrix $C=(c_{ij})$ of NxN size which represents an architecture of N nodes, where $c_{ij}$ indicates the presence or absence of a connection between the i and j nodes. It is possible to use $c_{ij}=1$ to indicate a connection and $c_{ij}=0$ to indicate an absence of connection. In fact, $c_{ij}$ could take real values instead of Booleans to represent the value of the connection weight between neuron "i" and "j", and in this way, architecture and connections can be developed simultaneously . These types of codification are generally very simple and easy to implement. However, they have a lot of disadvantages, such as scalability , the impossibility of codifying repeated structures, or permutation (i.e., different networks which are functionally equivalent can correspond with different genotypes) .

As a counterproposal to this type of direct codification method, there are also the indirect codification types in existence. With the objective of reducing the length of the genotypes, only some of the characteristics of the architecture are codified into the chromosome. Within this type of codification, there are various types of representation.

First, the parametric representations have to be mentioned. The network can be represented by a set of parameters such as the number of hidden layers, the number of connections between two layers, etc. There are several ways of codifying these parameters inside the chromosome . Although the parametric representations can reduce the length of the chromosome, the evolutionary algorithm makes a search in a limited space within the possible searchable space that represents all the possible architectures. Another type of non-direct codification is based on a representational system with the shape of grammatical rules . In this system, the network is represented by a set of rules which will build a matrix that represents the network. These rules have the shape of production rules, with antecedent and consequent part. This method has limitations in the way that, it needs to have a predefined number of steps in which the rules will be applied. It also does not allow the existence of recursive rules, and a compact genotypic representation (chromosome) does not imply a compact architecture.

Other types of codification, more inspired in the world of biology, are the ones known as "growing methods". With them, the genotype does not codify the network any longer, but instead of it, it contains a set of instructions. The decodification of the genotype consists of the execution of these instructions, which will provoke the construction of the phenotype . The instructions contained in the genotype are usually represented as tree structure, which means that the GP algorithm can be used in the evolution. During the decodification process, this tree is thoroughly covered, starting from the top and continuing through all its branches. The initial node represents the initial cell which, after the execution of the instructions, will give rise to the network. Each one of the nodes of the tree represents the operations which have to be applied to the corresponding cell and its two sub-trees specify the operations which will be applied to the child cells. For example, this process is taken as basic behaviour at  where the author is able to develop really complex ANNs for different proposes.

Another type of non-direct codification is based on the use of fractal subsets of a map . Fractal representation of the architectures is biologically more plausible than rule based representation. Three parameters are used which take real values to specify each node of the architecture: a border code, an entry coefficient and an exit coefficient. One application of this method can be found in .

One important characteristic is that, in general, these methods only develop architectures, which is the most common, or else architectures and weights together. The transfer function of each architecture node is assumed to have been previously determined by a human expert, and that it is the same for all of the network nodes (at least, for all of the nodes of the same layer), although the transfer function has been shown to have a great importance on the behaviour of the network . These methods have had little repercussion in the world of ANNs with EC. One of the few significant researches in which the transfer function is made evolve can be found in Dorado et al. . In it, a two-layer GA is used to design the architecture of an ANN and perform its training.

*2.3.2. Evolution of the learning rule*

Another interesting approach to the development of ANNs by means of EC is the evolution of the learning rule. This idea emerges because a training algorithm works differently when it is applied to networks with different architectures. In fact, and given that a priori, the expert usually has very few knowledge about a network, it is preferable to develop an automatic system to adapt the learning rule to the architecture and the problem to be resolved .

There are various approximations to the evolution of the learning rule , although most of them are based only on how the learning can modify or guide the evolution, and in the relation between the architecture and the connection weights. Actually, there are few works that focus on the evolution of the learning rule itself .

One of the most common approaches is based on setting the parameters of the BP algorithm: learning rate and momentum . Some authors propose methods in which an evolutionary process is used to find these parameters while leaving the architecture constant . Other authors, on the other hand, propose codifying these BP algorithm parameters together with the network architecture inside of the individuals of the population .

Due to the complexity involved in realizing a codification of all of the possible learning rules, it is necessary to establish certain restrictions to simplify this representation. Therefore, the search space will also be limited. Chalmers defined a learning rule as a lineal combination of four variables and six constants . Each individual of the population is a binary string which exponentially codifies ten coefficients and a scale value. From this work, others have emerged arriving at similar results , in which it is possible to see that the capacity of learning of an ANN can improve by means of evolution.

**3. MODEL**

As it was already said in section 1, the system proposed here mixes GP and GAs in order to develop and simplify ANNs. ANN simplification means to obtain a network with a small number of neurons and/or connections. GP is used to evolve full ANNs, i.e., it performs the evolution of the architectures and returns ANNs with the weights already set. Therefore, the networks returned by GP can be evaluated with the pattern sets. Therefore, the main evolutionary process is performed by GP, which is the one that evolves the ANNs.

However, GAs are also used to improve this evolutionary process. Their objective is to optimize the values of the connection weights of some networks. During the main evolutionary process (in which GP is used to evolve the networks), sometimes a GA is run in order to optimize the weights of some of the best networks found until that moment. This process is explained in more detail below.

*3.1. Evolution of ANNs by means of GP*

The development of an ANN by means of the use of GP is achieved thanks to the typing property of GP : each GP tree node has a type, and also, for those which aren't leaves of a tree (i.e., that have children) it will be necessary to establish the type of each one of their children. Therefore, with the objective of using GP to generate ANNs, the first thing to do is to define the types that are going to be used. These types are the following :

- TNET. This type identifies the network. It is only used in the root of the tree.
- TNEURON. This type identifies a node (or sub-tree) as a neuron, whether it is a hidden, an ouput or an input one.
- TREAL. This type identifies a node (or sub-tree) as a real value. It is used to indicate the value of the connection weights, i.e., a node having this type will be either a floating point constant.

With just these three types, it is now possible to build networks. However, the terminals and functions sets are more complicated. They contain the nodes of the tree, either with children (function set) or leaves (terminal set). The description is the following:

- ANN. It is the only node that will have a TNET type. As the tree must have the type TNET, this node will be the root of the tree, not being able to appear in another part. It has the same number of children as ouputs has the network. Each of these children has a TNEURON type because it will be a neuron.
- n-Neuron. Set of nodes which identify a neuron with n entries (other n neurons which pass their outputs to this one). These nodes will have TNEURON type. They will have 2*n children. The first n children of this node will designate the neurons or sub-networks that will be inputs to this neuron. The second n children will have the TREAL type, and they will contain the value of the corresponding input neuron connection weights (of the first n children) for this neuron.

- Input_Neuron_n. Set of nodes which define an input neuron that receives its activation value from the variable n. These nodes will be of a TNEURON type and they do not have children.
- Finally, and in order to have the values of the connection weights, random constants are added to the terminal set. The range of these constants is set below.

Table 1. Terminal and function sets for the GP system.

|  | Name | Type | Num. children | Children types |
|---|---|---|---|---|
| Function set | ANN | TNET | n | TNEURON, …, TNEURON |
|  | n-Neuron | TNEURON | 2*n | TNEURON, …, TNEURON, TREAL, …, TREAL |
| Terminal set | Input_Neuron_n | TNEURON | - | - |
|  | [-X, X] | TREAL | - | - |

Therefore, a neuron can be represented by two different nodes: n-Neuron for neurons that have n inputs and Input_Neuron_n for a neuron that receives as input the attribute n. A more formal description of these operations can be seen in Table 1.

In Fig 2, a simple network which can be constructed with this set of terminals and functions can be observed. In it, the nodes named "IN_x" correspond to "Input_Neuron_x". The network accepts 4 different inputs and it has two different outputs.

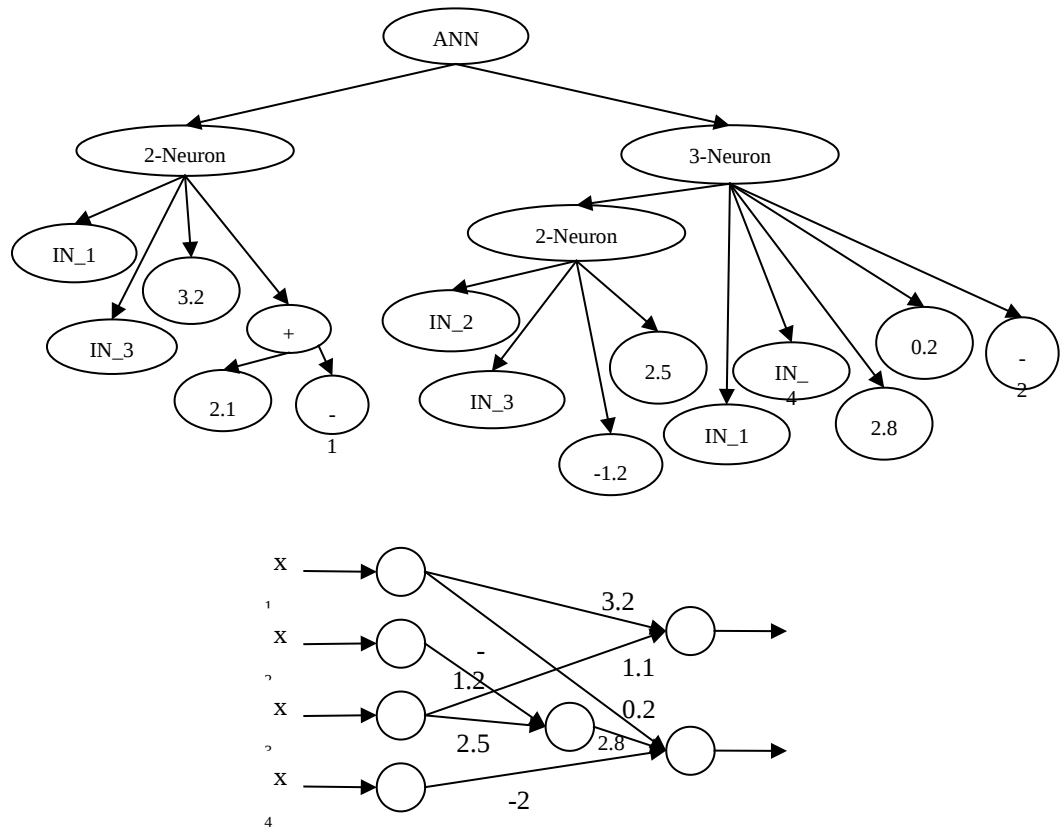

**Fig 2 GP tree and its correspondence ANN**

This typing system allows the construction of simple networks, but it has one great disadvantage: it does not allow the reutilization of a part of the network. With the set of specified operators, it is not possible for one neuron to give its output to more than one different neuron (except for network input neurons); i.e., the same neuron cannot be referenced various times from different parts of the ANN. This is a big drawback, because it eliminates one of the big advantages of ANNs, which is the reutilization of part of its structure. As ANNs have a high connectivity, they massively reutilize the results previously computed, converting many parts of the network into functional blocks.

In order to accomplish this, i.e., in order to be able to reference a neuron as an input of more than one processing element, the terminal and function sets have grown and the system has been extended to include a list which allows the referencing of previously used nodes. While the tree is being evaluated and the network is constructed, the neurons being added to the network are also stored in this list. The addition of neurons to the list is done automatically, at the same time that the tree is being evaluated. By means of a special operator, these neurons can be extracted from the list so they can be reutilized. This operator is a new node called "Pop", with a TNEURON type. When this operator is used as a node of the tree, instead of a new neuron it returns the neuron which is located at the beginning of the list. Therefore, a new neuron will not be created, but a reference (connection) to an existing one will be done.

However, it is desirable to return not only the neuron in the beginning of the list, but any neuron of the list. For this purpose, to extract neurons from the list with the objective of reutilizing them, an index will used for pointing out one of its elements (a neuron). When a "Pop" node is evaluated, the returned neuron will not be the one at the beginning of the list, but the one pointed by this index. A new node is needed to modify the position of this index. This operator is called "Forward" and has one child. When this operator is evaluated, it moves forward the index of the list and returns the result of evaluating its child.

Therefore, two new operators are needed: to extract neurons from the list and to modify the position of the index. The description of the new operators is the following:

- "Pop". TNEURON type. This node extracts the neuron from the list in the position pointed by the index. This node replaces the evaluation of a neuron because it returns one which already exists and, therefore, it does not have children.
- "Forward". TNEURON type. This node advances the list index in one unit. It has one child, with the TNEURON type. As opposed to the previous operator, its evaluation does not replace the evaluation of a neuron because its child will be a neuron and the evaluation of this node will have two effects: it advances the list index in one unit, and it returns the neuron resulting from the evaluation of its child.

```
    evaluate(node)
    begin

      case node of:

        ANN:
          begin
            create empty network, with input neurons
            create empty List
            Index = 1
            for i=1 to (number of outputs),
              neuron = evaluate(child i)
              set neuron as ith output of the network
            endfor
            return network
          end

        n-Neuron:
          begin
            create neuron
            for i=1 to n,
              input_neuron = evaluate(child i)
              input_weight = evaluate(child i+n)
              if (input_neuron is already input of neuron) then
                update weight of input_neuron with input_weight
              else
                set input_neuron as input of neuron with input_weight
              endif
            endfor
            add neuron to the List
            return neuron
          end

        Input-Neuron-n:
          Return input neuron n

        float f:
          return f

        Pop:
          return List[Index]

        Forward:
          begin
            Index = Index + 1
            neuron = return(child)
            return neuron
          end

      endcase
    end
```

Therefore, in a tree, it is possible to find, in the place of an "n-Neuron" node, a "Pop" node, meaning that instead of creating a neuron, a previously created one will be referenced. It is also possible to find a "Forward" node, which will increment the index of the list, but also it will return a neuron, whether it is a newly created one or one which already exists.

In the evaluation of the "n-Neuron" operator, therefore, it will be necessary to add this new neuron that is being created to the list. The evaluation of this operator implies the possible (and probable) creation of other neurons. The order of the addition of this neuron to the list with respect to the evaluation of the children of this operator and the creation of new neurons is crucial. There are two chances:

- The neuron is created and added to the list before its children have been evaluated and the respective neurons created. In this case, at the moment of evaluating a node with a TNEURON type in a sub-tree from one of this neuron's children, if this node to be evaluated is a "Pop" node, a neuron from the list will be referenced. Given that the preceding neuron is present on the list, it is possible that this neuron is the one referenced, which means that a recurrent link is being created. Therefore, this evaluation order allows for the creation of recurrent networks.

- The neuron is created, its children evaluated, and the node is added to the list after the evaluation of the children. In this case, in the creation of the links of the children neurons, this will not be present on the list, which means that recurrent connections to ancestor neurons will not be able to be realized. This is the study case of this work, in which recurrent networks are not being developed.

For a better understanding of the evolution of the set of terminals and functions, a pseudocode is included here to show how each one of the nodes is evaluated.

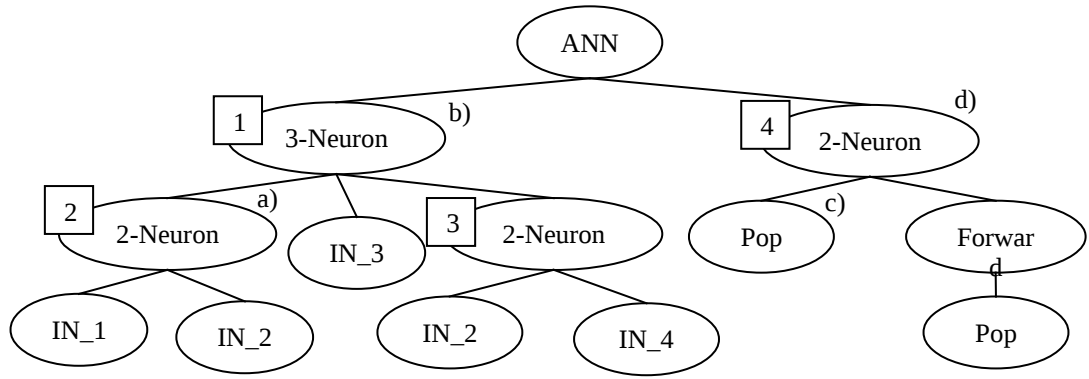

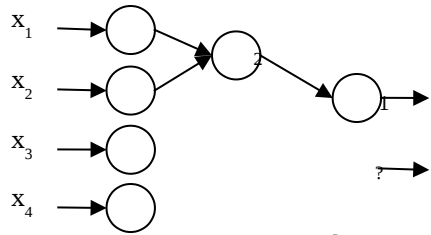

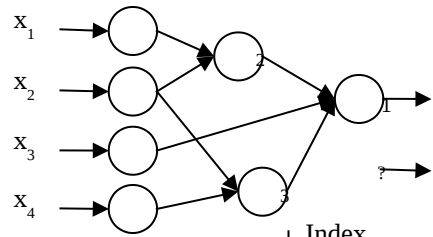

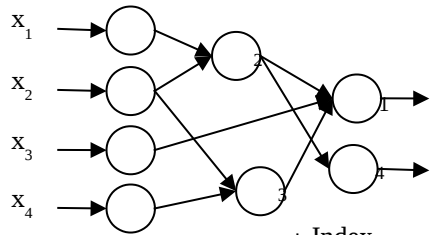

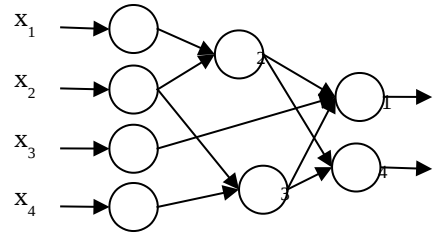

**Fig 3 GP tree with "Forward" and "Pop" nodes and its corresponding ANNs**

An example of a network which includes these operators can be seen in Fig 3. To simplify matters, in this figure, the constants which determine the connection weight values have been left out. As in the previous figure, the nodes denominated "IN_n" refer to "Input_Neuron_n". In this figure, it is possible to see the generating network in successive steps of the tree evaluation. The explanation of these steps, as they are labelled in the figure, is the following:

a) Network generated after the evaluation of the neuron labelled as "2". Two neurons have been created (labelled with numbers 1 and 2 inside squared boxes), but only one of them has been added to the list, because only that neuron has been completely evaluated, with all its children.

b) Network generated after the evaluation of the first "Neuron" node. Note that 3 neurons have been evaluated with their children, and they have been added to the list.

c) Network generated after the evaluation of the "Pop" node and the assignation of the neuron referenced in the list as input of the second output. Note that the node referenced by the "Pop" operator (i.e., the neuron pointed by the index in the list) is the one labelled with "2". This neuron is the first one inserted in the list because no "Forward" operators have been evaluated.

d) Network generated after the evaluation of the "2-Neuron" node. In this case, the evaluation of the "Forward" operator has made the index advance one position in the list. As consequence, the following "Pop" operator returns the neuron in the list labelled as "3".

It is important to keep in mind that, during the creation of a neuron, in the process of creating or referencing child neurons, a neuron can appear several times as input of another neuron. In this case, a new input connection from that neuron to the other is not established, but instead of it the existent connection weight is modified and the weight value of the new connection is added. Therefore, a common situation is that an "n-Neuron" operator is not referencing n different neurons, but instead of it is possible to have neurons repeated, especially if n has a high value. It is necessary to limit the n value, to set which is the maximum number of ancestors a neuron can have. A high value will surely provoke the described effect, i.e., that an effective use will not be made of all those inputs, but instead some inputs will be repeated. However, it is also necessary to take a high value in order to ensure a possibly high number of ancestors.

*3.2. Evaluation of the trees*

Once a tree has been evaluated, this genotype has been turned into a phenotype, i.e., a network with fixed weight values that can now be evaluated, meaning that it does not have to be trained. The evolutionary process requires the assigning of a fitness value to each genotype. In order to set this fitness value, a set of examples, called training set, is needed. Each example is composed by a set of inputs values and a desired output that the ANN should return from these inputs. This pattern set represents the problem, and this fitness value will be the result of the evaluation of the network with the pattern set that represents the problem. In this case, the result of this evaluation is the Mean Square Error (MSE) of the difference between the network outputs and the desired ones, on all of the patterns of the training set.

However, the fitness value is the result of the sum of two terms: this error value (ECM on the training set) and a penalty factor, introduced in order to make the system generate simple networks. This added value is a penalization value multiplied by the number of neurons of the network. In this way, and given that the evolutionary system has been designed to minimize a value error, if a value is added to the fitness value, it will cause a bigger network to obtain a worse fitness value, meaning that the appearance of simple networks is preferred because this added penalization value is proportional to the number of ANN neurons. The final fitness calculus used is the following:

$$fitness = MSE + N * P$$

where MSE is the mean square error of the network on the training pattern set, N is the number of neurons of the network, and P is the penalization value to the number of neurons. The P constant will have a low value, much less than the unit, and, as will be showed in the experiments, it is crucial in the evolution of the system. In this way, the system will try to evolve simple systems with few neurons.

A common problem when generating and training ANNs is the tendency to overfit them. This can be observed when the training fitness value (in this case, MSE) keeps improving during the training process while the test value begins to worsen. In that case, the network is losing the capacity of generalization. In order to avoid this problem, in addition to the training set, a different pattern set is also used, with the objective of making a validation. This validation set controls the training process in order to avoid the overtraining of the networks . By means of the use of this validation set, the training happens in a similar way, but this time each network returned by the evolutionary process is evaluated in the validation set, which gives a test fitness estimation. The system always returns the network with the best results obtained in the validation set, even if the training process keeps going and other networks with better training results are obtained. This is done in a similar way to the process realized by the technique of early stopping . In this way, the validation set provides an estimation of how a network is going to behave in the test done with it.

*3.3. Optimization of the connection weights by means of a GA*

The whole process described can itself perform the evolution of ANNs with no need of any other process, i.e., the networks returned have their weights already set. However, this process can be improved. In the generation of ANNs by means of GP, the creation of the constants that represent the connection weights is done at random at the beginning. New constants are created only on a mutation operation performed on a tree, and exchanged between trees on a crossover operation among two trees. Therefore, the constants generated at the beginning of the evolutionary process (on the creation of the initial trees) have a very high importance in the whole evolutionary process, since the only way of generating new constants (by means of the mutation operator) is done with a very low probability.

This is the reason why a constant optimization process is run in order to optimize the values of the constants of the individuals. Not only new constants are generated, but they are also optimized. This process is done by means of a GA. Traditionally, ANNs are trained by means of the Backpropagation algorithm. However, this algorithm cannot be used with the networks generated with this system because they are not classically distributed into layers, being totally connected from one layer to the next one. For this reason, a GA is used to train the ANN.

The optimization of the connection weights is done by means of a generic constant optimization process . This process is valid for the optimization of any GP expression that uses constants. In this case, those constants are used to represent the connection weights. This optimization process is run to optimize a small percent ($F_{sel}$) of GP individuals, after the execution of a certain number of generations ($N_{int}$). The individuals to be optimized are the best ones, with the exception that no repeated individuals are to be optimized. This optimization process is run after the creation of the first GP population, and after the execution of $N_{int}$ GP generations. Using this constant optimization process, the whole evolutionary process is the following:

1. Create the initial GP population
2. Execute the constant optimization system to optimize the $F_{sel}$ best individuals
3. Run $N_{int}$ generations of the GP algorithm until the termination criterion is fulfilled
4. If the termination criterion is satisfied, end the execution. Otherwise, go to step 2

The optimization process is based on the execution of a different GA with a real value codification for each of the GP individuals to be optimized. Therefore, for a single GP individual a different GA is run. In that GA each of the genes represents one weight of the network (one constant in the GP genotype). This GA is run and, as consequence of this, an evolutionary process is generated in which the network's weights are adjusted. After the execution of this process, the constants of the GP individual to be optimized will adopt those values of the best GA individual found and, consequently, the fitness value of that GP individual is also updated (with a better value).

When running a GA optimizing a particular GP individual, the evaluation of each GA individual to calculate its fitness function is similar to the evaluation of the GP individuals. In the case of evaluation of a GA individual, the constants are taken and are assigned as the weights of the connections of the network that is being optimized. After this, the network is evaluated with the pattern sets. Therefore, the fitness value is the same as the one returned in the GP evolution.

More specifically, the constant optimization process applied to a particular GP individual (that represents a network) is the following:

1. A GA is created. The chromosome length will be the number of weights of the network represented by the GP individual. Each of the genes represents a weight.
2. The population of the GA is initialized with $N_{ga}$ individuals. This is done in the following way:
    - ✓ As the GP individual already has weight values, a GA individual is created with these initial values and the already existing fitness value of the GP individual.

- The remaining $N_{ga}$-1 individuals of the population are randomly generated and evaluated. The fitness function is the one previously explained, i.e, set the values of the genes of the chromosome as the connection weights and return as fitness the result of evaluating the network with the pattern sets.
3. Run the GA until the execution of $N_{gagen}$ generations or until there is no change in the best fitness in 5 generations.
4. Once the execution of the GA is finished, the best individual is taken and the values of the connection weights in the GP individual are updated with those values of the chromosome of the best GA individual. Therefore, this optimization process can be seen as a Lamarckian strategy . The fitness value of the GP individual is also changed for the fitness value of the best GA individual.

Once this optimization process has finished, the GP evolution keeps going on. After the execution of other $N_{int}$ GP generations, the optimization process is repeated again with the $F_{sel}$ best individuals of the GP population.

There are 5 parameters that rule this optimization process:
- $I_{ga}$: Interval ($[-I_{ga}, I_{ga}]$) in which the weights can take values, i.e., this parameter sets the upper and lower of all of the genes in the genotype.
- $N_{int}$: Number of GP generations that must be run between 2 executions of the optimization process.
- $F_{sel}$: Percent of GP individuals to which the optimization process is applied.
- $N_{ga}$: GA population size.
- $N_{gagen}$: Maximum number of GA generations. The GA is stopped after the execution of $N_{gagen}$ generations of after 5 generations without any improvement in the best individual. It has been observed that this parameter hardly has any influence on the overall process, since most of the times the $N_{gagen}$ generations are not reached because it does not usually happen that 5 generations with no improvement in the best individual are done.

# 5. RESULTS

## 5.1. Problem description

In order to test the behaviour of this system, experiments were done with diverse problems of different natures. The authors has selected 6 problems where 3 were seleted (Apendicitis, Breast Cancer and Heart Diase) because are commonly used biomedical problems to test different classification and data mining tools. The other 3 problems (Iris Flower, Mushroom and Ionosphere) were selected because are commonly used problems to compare the performance among ANNs techniques and they were incorporated in order to obtain a more general overview of the proposed method. Table 4 shows a small summary of the most important characteristics of the 6 problems to be solved. Specifically, the problems used here are the one that are in the well-known UCI public repository for Machine Learning. It is important to keep in mind that for the most of them, to have only one output is desired. However, in the problem of Iris the objective is to do a classification in three possible cases, meaning that networks are generated with 3 possible outputs, one for each case.

Table 4. Summary of the most important features of the problems to be solved

|  | Num. inputs | Num. data points | Num. outputs |
|---|---|---|---|
| Appendicitis | 8 | 106 | 1 |
| Breast Cancer | 9 | 699 | 1 |
| Iris Flower | 4 | 150 | 3 |
| Mushroom | 22 | 5644 | 1 |
| Heart Disease | 13 | 303 | 1 |
| Ionosphere | 34 | 351 | 1 |

The attributes of all these databases have been normalized between 0 and 1 and the databases were randomly divided into two parts, taking 70% of the data as the training set and the other 30% as the validation set.

*5.2. Parameter values*

For each parameter, values were found as a result of a preliminary empirical analysis which seem to make the system behave robustly. Table 6 shows a summary of the parameters studied here and those intervals of the parameters studied here. Also, for each parameter, the chosen value for the experiments of the next section is shown in this table.

Table 6. Values used in each parameter

| | Parameter | Value |
|---|---|---|
| **GP Parameters** | Population size | 1000 |
| | Crossover rate | 95% |
| | Mutation probabilitiy | 4% |
| | Selection algorithm | 2-individual tournament |
| | Creation algorithm | Ramped Half&Half |
| **ANN generation parameters** | Maximum number of inputs to each neuron | 5 |
| | Maximum height | 6 |
| | Penalization | 0.00001 |
| **GA optimization parameters** | $I_{ga}$ | 20 |
| | $N_{int}$ | 80 |
| | $F_{sel}$ | 10 |
| | $N_{ga}$ | 50 |
| | $N_{gagen}$ | 50 |

*5.3. Methodology*

The system porposed here has been compared with other ANN generation and training methods in order to evaluate its performance.

When comparing classification algorithms, the most used method is cross validation for estimating the accuracy of the algorithms, and then using t-tests to confirm if the results are significantly different. In the cross validation method, the data set D is divided into k sets $D_1$, …, $D_k$ that do not overlap (*k-fold cross validation*). In each i iteration (that varies from 1 to k), the algorithm trains with the D\$D_i$ set and a test is carried out in $D_i$. However, some studies have shown that the comparison of algorithms using these t-tests in cross validation leads to what is known as type I error .

In the performance of the *k-fold cross* validation method was analyzed, combined with the use of a t test. That work proposed to modify the used statistic and it was justified that it is more effective realizing k/2 executions of a *2-fold cross validation* test, with different permutations of the data, than realizing a *k-fold cross validation* test. As a solution between test accuracy and calculation time, it was proposed to realize 5 executions of a cross validation test with k=2, resulting in the name 5x2cv. In each one of the 5 iterations, the data is divided randomly into two halves. One half is taken as input of the algorithm, and the other one is used to test the final solution, meaning that there are 10 different tests (5 iterations, 2 results for each one) .

In  the 5x2cv method is used to compare different techniques based on evolutionary methods in order to generate and train ANNs. In that work, the results that are presented are the arithmetic means of the accuracies obtained in each one of the 10 results generated by this method. These values are taken as basis for comparing the technique described in this work with other well-known ones. Each one of these 10 test values were obtained after the training with the method described in this work.

An additional problem of the 5x2cv technique for comparing methods is that this technique requires the division of the pattern set into two halves. This may not be problematic when working with sufficiently large sets. However, here as well as in the work in which the comparison is based, the pattern sets that are used are quite small (see Table 2, problems of breast cancer, iris, heart disease and ionosphere). To avoid overfitting, this work proposes to split the training set into training and validation sets order to perform a validation of the obtained networks and to take the best results from this validation subset. In case the 5x2cv is used, this involves dividing the training set, which is half of the initial pattern set, into two parts. This second partitioning provokes that either the training or the validation is produced with a very reduced number of data points, which will surely not be representative of the search space that is being explored. To verify this effect, different experiments have been realized. In these experiments, the training set has been divided into two parts (training and validation) extracting a total of 50%, 40%, 30%, 20%, 10%, and 0% of the training data points to the validation set. As was previously explained, this validation process makes the system return the network that has produced the best results in the validation set (which is an estimate of the results that will be offered in the test). In case of extracting a 0%, which means that no validation is being realized, the results will present the effect of overtraining. The results are shown in Table 7, which show the accuracies obtained in the tests up to a maximum of 500000 fitness function executions.

Table 7: Comparison of the accuracies obtained in the 5x2cv test leaving different amounts of the training pattern set to perform validation with a n effort of 500000.

| Validation Percent | Accuracy | | | |
|---|---|---|---|---|
| | Breast Cancer | Heart Cleveland | Iris | Ionosphere |
| 0% | 96.03 % | 79.96 % | 95.21 % | 89.70 % |
| 10% | 93.28 % | 70.91 % | 89.22 % | 76.25 % |
| 20% | 94.18 % | 75.42 % | 93.29 % | 84.16 % |
| 30% | 95.25 % | 78.50 % | 94.66 % | 87.02 % |
| 40% | 95.56 % | 78.42 % | 95.05 % | 87.21 % |
| 50% | 95.57 % | 79.35 % | 95.21 % | 87.43 % |

Table 7 shows that the best results are obtained when not splitting the training set another time into training and validation, i.e., without using any validation set. This means that when the training is partitioned into validation and test, these new sets are not sufficiently representative of the search space, especially the validation one. The fact that the bigger the validation set is, the better the obtained results are, leads to thinking that it does not have the sufficient size to be representative of the pattern set and be able to offer a reliable estimation of the test. In case of having a sufficiently large pattern set, it is recommended to divide it into three parts: training, testing and validation. However, if this is not possible, it should be divided into only training and testing.

A more detailed description of all the algorithms with which this technique is compared can be found in . In this work, the average times needed to achieve the aforementioned results is indicated. Given that it is not possible to be able to use the same processor in order to make the comparison, the approximate computational effort needed to achieve those results was calculated. This effort is measured as the number of times that the pattern file is evaluated. The computational effort for each technique can be measured on the basis of population size, the number of generations, the number of times that the BP algorithm is applied, etc. In general, the calculation varies for each algorithm that is used. Despite that, the calculation is very similar for all of them, and it is based on realizing the calculation of the effort needed in evaluating each individual (which could imply evaluating an ANN only once, or training an entire ANN, or something in between), and multiplying that value by the population size and the number of generations.

For each comparative table that is shown (Tables 9, 10, 11, 12 and 13), each square corresponds to a particular problem with a particular technique. Three different values are shown in them. On the left, the accuracy value obtained in can be seen. Below, the computational effort needed to obtain that value with that particular technique is shown. On the right side, the average value of the obtained accuracy with the technique described here, corresponding to the average of the results obtained with that computational effort value. If the computational effort needed for each technique is lower to 2,000,000 fitness function executions, the accuracy value shown by the technique described in this work will be the one which corresponds to that effort. However, if it is greater, the accuracy value shown will correspond to the one obtained after 2,000,000 fitness function executions. However, this value shown is not the result of a single execution, but the average of the different independent runs of the cross-validation algorithm.

These techniques with which the comparison will be made can be divided into three groups: training of ANNs by means of evolutionary algorithms, feature selection, and design and training of ANNs by means of evolutionary algorithms.

*5.4. Training of ANNs by means of evolutionary algorithms*

The first group of techniques used for comparison are the ones that only use genetic algorithms to train ANNs with an already fixed topology. Table 8 shows a summary of the network topologies used in together with an arithmetic mean of the resultant network typologies using the method proposed here. It is important to keep in mind that even though results in the cited work were obtained using ANNs with a totally connected hidden layer and a determined number of neurons in the hidden layer, the network topologies presented according to the method proposed in this study do not correspond to a classic topology divided into layers, because the hidden neurons can have any kind of interconnection between themselves and the input and output ones with the possibility of also having connections between input and output neurons. Even the topologies found in the method proposed here have a slightly higher number of neurons than the ones used in , the number of connections of the networks found is much lower.

Table 8. Network topologies used for the comparisons.

| Proposed here | Cantú-Paz 2005 |
|---|---|

|  | Num. inputs | Num. hidden neurons | Num. outputs | Num. connections | Num. inputs | Num. hidden neurons | Num. outputs | Epochs | Num. connections (totally connected) |
|---|---|---|---|---|---|---|---|---|---|
| Breast cancer | 9 | 5.55 | 1 | 23.6 | 9 | 5 | 1 | 20 | 50 |
| Iris Flower | 4 | 11.40 | 3 | 40.75 | 4 | 5 | 3 | 80 | 45 |
| Heart Cleveland | 13 | 5.05 | 1 | 20.7 | 26 | 5 | 1 | 40 | 135 |
| Ionosphere | 34 | 10.45 | 1 | 43.85 | 34 | 10 | 1 | 40 | 350 |

Table 9 shows the results obtained by the method proposed here in comparison with the ones obtained with the traditional *backpropagation* (BP) algorithm and trained by means of GA, with either binary or real codification (algorithm G3PCX ).

When using binary codification, the configuration of the genetic algorithm was the following:

- Chromosome of 16 bits per weight in the [-1, 1] range.
- Population size of $n = \lfloor 3\sqrt{l} \rfloor$, where l is the number of chromosome bits.
- Mutation rate of 1/l.
- Two-individual tournament without replacements.
- In each generation the whole population was replaced without elitism.
- After 100 generations the algorithm was stopped and the result was the individual with highest accuracy.

On the other hand, the genetic algorithm with real codification, denominated as G3PCX , presents a similar configuration with the following variants:

- Chromosome length: l, where l is the number of weights.
- Population size: $n = \lfloor 30\sqrt{l} \rfloor$, where l is the number of weights.
- The algorithm was stopped after n iterations without improvement in the best solution, or after 50n executions of the fitness function.
- Uses a *steady-state* algorithm, i.e., only one population that evolves and is not replaced by another result from applying genetic operators to it.

Table 9. Comparison of results from ANN training methods

|  | BP | G3PCX | Binary |
|---|---|---|---|
| Breast cancer | 96.39 | 98.94  95.90  2247200 | 98.88  95.74  8400 |
| Iris Flower | 94.53 | 89.73  95.11  1566450 | 88.67  85.23  7000 |
| Heart Cleveland | 78.17 | 90.42  79.85  6055200 | 87.72  77.44  13900 |
| Ionosphere | 84.77 | 64.10  90.05  15736050 | 74.10  83.25  22400 |
| Mean | 88.46 | 85.79  90.22 | 87.34  85.41 |

By using lamarckian and baldwinian strategies, results have also been taken from other training processes by means of a GA (with binary codification), but this time refining the weights in the genetic algorithm by means of the use of the BP algorithm. In the case of the lamarckian strategy, the BP algorithm is applied to the chromosomes of the individuals during the evaluation of the fitness function in a certain percentage of the individuals of the population, and the changes done to the chromosomes (weights of the ANNs) are permanent. The resulting fitness value will be the result of applying the BP algorithm. On the other hand, the baldwinian strategy is similar to the lamarckian one, with the difference that the changes in the individuals after the execution of the BP algorithm are not permanent, the BP algorithm is only executed to calculate the fitness value. In addition, in these experiments, the best set of weights (i.e., the best chromosome) found in the baldwinian strategy was used to train a network by means of the BP algorithm. Table 10 shows the comparison between the method proposed in this work and the lamarckian and baldwinian strategies, applying these strategies to 5% of the population. For each strategy, the number of BP algorithm iterations is indicated (1, 2, and 5 iterations in the cases of 1BP, 2BP, and 5BP respectively). Table 11 shows a similar comparison, but this time these strategies were applied to 100% of the population. The genetic algorithms used in these experiments have the same parameters as the binary GA used in the previous case.

Table 10. Results of the comparisons with evolutionary training strategies, using lamarckian and baldwinian strategies in 5% of the population

| Baldwinian | Lamarckian |
|---|---|

|  | 1BP | 2BP | 5BP | 1BP | 2BP | 5BP |
|---|---|---|---|---|---|---|
| Breast cáncer | 98.48  95.74 8820 | 98.91  95.82 9240 | 99.03  95.82 10500 | 98.88  95.74 8820 | 98.74  95.82 9240 | 99.08  95.82 10500 |
| Iris | 89.47  85.50 7350 | 91.07  85.65 7700 | 87.20  85.91 8750 | 89.20  85.50 7350 | 88.00  85.65 7700 | 88.13  85.91 8750 |
| Heart Cleveland | 88.68  77.49 14595 | 89.25  77.51 15290 | 89.21  77.53 17375 | 86.45  77.49 14595 | 88.82  77.51 15290 | 87.98  77.53 17375 |
| Ionosphere | 68.43  83.35 23520 | 69.43  83.42 24640 | 67.86  83.90 28000 | 65.12  83.35 23520 | 65.88  83.42 24640 | 64.65  83.90 28000 |
| Mean | 86.26  85.52 | 87.16  85.60 | 84.89  86.04 | 84.91  85.52 | 85.36  85.60 | 84.96  86.04 |

Table 11. Results of the comparisons with evolutionary training strategies, using lamarckian and baldwinian strategies in 100% of the population

|  | Baldwinian | | | Lamarckian | | |
|---|---|---|---|---|---|---|
|  | 1BP | 2BP | 5BP | 1BP | 2BP | 5BP |
| Breast cáncer | 98.83  95.92 16800 | 98.86  95.97 25200 | 98.60  96.12 50400 | 98.88  95.92 16800 | 98.94  95.97 25200 | 98.86  96.12 50400 |
| Iris Flower | 91.33  86.83 14000 | 89.87  87.41 21000 | 91.07  92.08 42000 | 92.40  86.83 14000 | 93.20  87.41 21000 | 92.00  92.08 42000 |
| Heart Cleveland | 88.58  78.25 27800 | 88.45  79.40 41700 | 88.32  80.59 83400 | 86.87  78.25 27800 | 88.98  79.40 41700 | 87.66  80.59 83400 |
| Ionosphere | 73.88  86.78 44800 | 73.15  87.93 67200 | 72.89  88.93 134400 | 74.25  86.78 44800 | 74.03  87.93 67200 | 73.77  88.93 134400 |
| Mean | 88.15  86.94 | 87.58  85.54 | 87.72  89.43 | 88.1  86.94 | 88.78  85.54 | 88.07  89.43 |

*5.5. Feature selection*

The next technique which is used in to realize comparisons is based on the selection of variables of the problem . This technique is based on having a genetic algorithm with binary codification in which each bit inside of the chromosome will indicate if a determined variable will or will not be used for the training. The evaluation of the individuals, therefore, will consist of training the ANN (which will have a fixed structure described in [89]) with the variables fixed for the chromosome of the individual. The population was randomly initialized with $\lfloor 3\sqrt{l} \rfloor$ individuals, but with a minimum size of 20. A standard crossover with a probability of 1.0 was used and a mutation rate of 1/l. The networks were designed and trained in accordance with Table 8. The algorithm was stopped either when no change was found in the best solution for five generations or after arriving at a limit of 50 generations. However, this limit was never achieved because the algorithm found a good solution much before, i.e., the 5 generation limit without any changed was reached before. Table 12 shows a comparison of this technique with the one presented in this work.

Table 12. Results of the comparison with the variable selection method

|  | Feature selection |  |
|---|---|---|
| Breast cancer | 96.48 | 95.93 |
|  | 20000 |  |
| Iris Flower | 93.60 | 95.10 |
|  | 80000 |  |
| Heart Cleveland | 84.72 | 79.29 |
|  | 40000 |  |
| Ionosphere | 87.00 | 85.96 |
|  | 40000 |  |
| Mean | 90.45 | 89.07 |

5.6. Design of ANNs by means of evolutionary algorithms

In conclusion, the next set of techniques with which this work is compared to refers to the use of evolutionary algorithms to design neural networks. The techniques to be compared with are the following:

- Connectivity Matrix.
- Pruning.
- Finding network parameters.
- Graph-rewriting grammar.

In all these techniques, in order to evaluate the accuracy of each network generated by any of these methods, 5 iterations (independent runs) of a 5-*fold* cross validation test are conducted, which have a notable influence on the computational effort needed to achieve the results presented.

The connectivity matrix technique is based on representing the topology of a network as a binary matrix: the element (i,j) of the matrix will have a value of 1 if a connection exists between i and j, and zero if there is no connection. A genetic algorithm with binary codification can be easily used because the chromosome is easily obtained by linking the matrix rows together . In this case, the number of hidden neurons indicated in Table 8 is used, and connections have been allowed between inputs and outputs, meaning that the length of the chromosomes is l =

(hidden + outputs)*inputs + hidden*outputs. A multipoint crossover was used with a probability of 1.0 with l/10 crossover points and a mutation rate of 1/l. The population had a size of $\lfloor 3\sqrt{l} \rfloor$ individuals with a minimum of 20. The algorithm was stopped after 5 generations without improving on the best solution or if a maximum of 50 generations was achieved. Each network was trained with the Backpropagation algorithm.

The pruning technique is based on a representation similar to the previous one. However, the method is different. It begins with a totally connected network , which is trained by means of the BP algorithm according to the parameters in Table 8. When this network is obtained, the evolutionary algorithm is executed. The evaluation function of each individual will consist of taking the previously trained network and eliminating those weights whose value in the connectivity matrix is equal to 0, in order to evaluate it afterwards with the training set, without further training. The networks begun with the topologies shown in Table 8, with the same configuration of parameters as in the previous case.

The finding of the network parameters is a different approach because in this case an evolutionary algorithm is used to find the general designing and training parameters of the networks . In this case, these parameters are the number of hidden neurons, the BP algorithm parameters, and the initial range of the weights. The chromosome's longitude was 36 bits, divided in the following way:

- 5 bits for the learning rate and the coefficient β of the activation function, in the [0, 1] range.
- 5 bits for the number of hidden neurons, in the [0, 31] range.
- 6 bits for the number of BP epochs.
- 20 bits for the upper and lower values of the initial weights range (10 bits for each value), and their ranges were [-10, 0] and [0, 10] respectively.

The evaluation of an individual consists on the construction of the network, its initialization and its training according to the parameters. The population had 25 individuals and was randomly initialized. The algorithm used a two-point crossover with a probability of 1.0 and a mutation rate of 0.04. As in the rest of the experiments, a two-individual tournament selection algorithm without replacements was used and the execution was stopped after 5 generations with no change in the best solution or after having reached a limit of 50 generations.

Finally, the graph-rewriting grammar consists on a connectivity matrix which represents the network. As opposed to the previous cases, the matrix is not codified directly in the chromosome, but instead a grammar is used to generate the matrix. The chromosome only has rules which convert each element of the matrix into sub-matrixes of 2x2. In this grammar, there are 16 terminal symbols that are matrices of 2x2, and 16 non-terminal symbols. The rules have the n→m form, where n is one of the non-terminal symbols, and m is a non-terminal 2x2 matrix. There is a starting symbol and the number of steps is determined by the user.

The chromosome contains the 16 rules in the following manner: it contains the 16 right sides of the rules because the left side is implicit in the position of the rule. To evaluate the fitness of an individual, the rules are decodified and a connectivity matrix is constructed by means of the same rules. The network, which is trained by means of BP, is generated from this matrix.

For the application in these problems, the number of steps is limited to 8, meaning that the results are networks with a maximum number of 256 elements. The size of the chromosome, therefore, is 256 bits (4 2x2 binary matrices for each one of the 16 rules). The population size is 64 individuals, with a multi-point crossover with a probability of 1.0 and l/10 crossover points and a mutation rate of 0.004. The algorithm was stopped after 5 generations with no improvement in the best individual or after 50 generations.

The results obtained with these 4 methods, in comparison with the method described in this work, can be seen in Table 13. As in the previous tables, for each problem and each comparison done, each cell shows in the right the value obtained when this method was stopped when the computational effort reached the number in the bottom. Therefore, the differences in comparing this method to the other are the number fitness functions allowed to be run the algorithm.

Table 13. Results of the comparisons using diverse network design methods

|  | Matrix | Pruning | Parameters | Grammar |
|---|---|---|---|---|
| Breast cancer | 96.77  96.20 92000 | 96.31  95.60 4620 | 96.69  96.16 100000 | 96.71  96.07 300000 |
| Iris Flower | 92.40  95.25 320000 | 92.40  83.65 4080 | 91.73  95.25 400000 | 92.93  95.10 1200000 |
| Heart | 76.78  80.02 | 89.50  76.90 | 65.89  80.01 | 72.8  80.01 |
| Cleveland | 304000 | 7640 | 200000 | 600000 |
| Ionosphere | 87.06  89.62 464000 | 83.66  82.57 11640 | 85.58  89.24 200000 | 88.03  89.73 600000 |
| Mean | 88.25  90.27 | 90.46  84.68 | 84.97  90.16 | 87.61  90.22 |

**6. DISCUSSION**

As the tables show, the results obtained by the method proposed here are in the same order as those presented in , improving them most of the time. However, these good results are only one of the features of this system. This section described these features.

*6.1. Results*

Section 5.4 show a comparison of the results obtained by this method and other ANN training methods by means of EC and hybrid techniques. The accuracy values obtained by the technique described here in the 5x2cv tests demonstrate that the results offered are similar to the ones attained using other tools, improving them in many cases, especially in those in which a lot of computational capacity is required (such as in hybrid techniques like, for example, lamarckian strategies). However, the techniques to which it is compared begin from a fixed network topology, meaning that it is still necessary to have the intervention of a human expert in those cases. The tool described here, on the other hand, and as has been demonstrated, is capable of offering even better results without the need for that kind of human intervention.

Moreover, section 5.6 reveals a comparison with a set of techniques that do not need a predetermined architecture beforehand. In other words, no longer is an expert's intervention needed in order to determine the ANN topology and connectivity. These techniques, because of joining the architectural evolution with the training of the weights, require an enormous computational load. Table 13 shows that in only one of the comparisons, the technique described here offers worse results. In the remainder, the accuracies achieved show much better results than those offered using other techniques.

Tests have shown that the cost necessary to obtain the results, in many cases, is less that show in Table 13. In some problems (breast cancer, heart disease) there is a fall in the test accuracy after an initial improvement. This is caused by the overfitting of the networks. This problem, as was already explained, is caused because there is not validation set. If larger data sets were used, the training set could have been splitted again into training and validation for avoiding this effect.

On the opposite side, in other problems the accuracy value keeps improving, so when the maximum number of fitness function evaluations is reached, the system has no achieved the maximum accuracy value that can obtain for that problem.

Therefore, the accuracy values shown on the tables are not the best that this technique can offer. In some cases because the stopping of the algorithm is too early, the results would improve if the algorithm kept running. In other cases, because better values were previously reached and would be maintained if a validation set was used. If this had happened, it is expected that the test accuracy does not have big falls, and the resulting test accuracy would be very similar to the best test accuracy found during the training. A comparison of these best test accuracies with the accuracies obtained with other tools can be seen on table 14, ordered by the accuracy value.

Table 14. Comparison of the best results found

|       |   | Breast cancer     |        | Iris Flower    |        | Heart Cleveland   |        | Ionosphere       |        |
|-------|---|-------------------|--------|----------------|--------|-------------------|--------|------------------|--------|
| Best  | 1 | Matrix            | 96.77% | Proposed here  | 95.28% | Pruning           | 89.50% | Proposed here    | 90.09% |
|       | 2 | Grammar           | 96.71% | Grammar        | 92.93% | Proposed here     | 80.69% | Grammar          | 88.03% |
|       | 3 | Parameter         | 96.69% | Matrix         | 92.40% | Matrix            | 76.78% | Matrix           | 87.06% |
|       | 4 | Pruning           | 96.31% | Pruning        | 92.40% | Grammar           | 72.80% | Parameter        | 85.58% |
| Worst | 5 | Proposed here     | 96.21% | Parameter      | 91.73% | Parameter         | 65.89% | Pruning          | 83.66% |

As can be seen on this table, the results obtained by the system described here a better than the ones returned by the rest of the techniques in most of the problems, except in the case of breast cancer. However, in this problem the difference in the accuracies obtained is very low, there is only a variability of the 0.56% between the results obtained by the tool described here and the one that has returned the best result. In other problem, heart disease, the technique proposed here gave the second best results, behind pruning. However, this latter technique only has offered good results in that particular problem, with very low accuracies in the rest of the problems, and therefore it is a very unstable tool.

A mean of the position of each technique on table 14 can be done. With these values, a quantification of the goodness of each tool can be performed. With this measure, these tools can be ordered as can be seen on table 15.

This table shows that the technique described here has the first place in accuracy. Therefore, it can be concluded that the results that returns are the best in average independently of the problem to solve.

Table 15. Techniques ordered by accuracy

|  |  | Technique | Average position table 13 |
|---|---|---|---|
| Best | 1 | Proposed here | 2.25 |
|  | 2 | Grammar | 2.5 |
|  | 3 | Matriz | 2.5 |
|  | 4 | Pruning | 3.5 |
| Worst | 5 | Parameter | 4.25 |

*6.2. Independecy with the expert*

As was already explained, one of the main goals of this work was to develop a system in which the expert has a minimum participation, to eliminate the excessive effort that has to do in the classical ANN development process. The development of this system has lead to the appearing of some parameters.

One could think that the fact of having to set the values of those parameters eliminates that desired expert independency. In section 5.2, a value is taken for each parameter and is used for the comparison with other tools in the solving of problems with different complexity because previous experiments have show that the algorithm is robust in a certain interval with few influence of the exact value. Therefore, by using these parameter values (or other values inside these ranges), the parameters do not have to be set, and thus the effort that the expert has to do is minimal. This effort is referred to the classical actions in machine learning, data analysis and preprocessing.

Moreover, the rest of the techniques used here in the comparison need some participation from the expert. The ones used in section 5.4 need a previous network design. The tools used in section 5.6 automatically perform this design task, but all of them still require some effort by the expert, who must perform some tasks like the design of an initial network (in the case of pruning) or establish how can be the networks that will be developed (in the case of parameter search or graph-rewriting grammar). This makes these tools not to be completely independent from the expert, and this expert still has to make some effort in order to correctly apply them. This does not happen with the technique described in this work.

*6.3. Discrimination of the input variables*

Another important advantage of this technique is that it allows the discrimination of those variables that are not important for solving the problem. This is done by the evolutionary process, because in the resulting ANN, the variables which are not significant for resolving the problem do not appear in the network. This feature selection can be very useful to give insight to the problem domain . In addition, the obtained network does not have the architectural limitation of having to be separated into layers, but rather, it can eventually have any type of connectivity.

The reduction of the number of input features can be seen on table 16, which shows the average number of variables used by different networks when 500.000 fitness function evaluations were performed. The initial number of input features is also shown. As can be seen, the number of inputs is very low in comparison with the original number, most of all in problems with many features, like ionosphere.

Table 16. Comparison of the features used in each problem

|  | Num. features | Used features | PCA | | | | |
| --- | --- | --- | --- | --- | --- | --- | --- |
|  |  |  | 1% | 2% | 5% | 10% | 15% |
| Breast cancer | 9 | 6.15 | 8 | 8 | 7 | 3 | 3 |
| Iris Flower | 4 | 3.55 | 3 | 3 | 2 | 2 | 1 |
| Heart Disease | 13 | 6.95 | 11 | 9 | 7 | 4 | 2 |
| Ionosphere | 34 | 10.175 | 15 | 7 | 3 | 2 | 1 |

This table also shows a principal component analysis (PCA) of these problems. This analysis is based on the study of the database from the point of view of the input features, because they are usually quite correlated. Therefore, there is usually much redundant input information. The PCA computes the eigenvectors and eigenvalues of the covariance matrix, and thus reduces the dimensionality of the input features. As result, this algorithm returns a set of features (a lower number than the original) which are orthogonal and, therefore, incorrelated. The importance of each variable in PCA is given by its variance. To perform a PCA analysis one must indicate which percent of the total variance is desired the new model to explain, i.e, the new feature set and the transformations performed to reach it.

Table 16 shows different values of the number of variables for different values of variance percent and problems. These values represent the number of features of the model resulting after eliminating from it those variables that contribute less than that percent to the total variance of the data set. As can be seen on the table, the number of variable used for each problem corresponds to the resulting number of applying a PCA with a very low variance percent, lower than 10%, and sometimes lower than 5% (in the case of iris flower problem) or even than 2% (in the case of ionosphere problem). The system described in this work discriminates a set of variables that contain a great amount of information of the original database, with the additional feature from PCA that no transformation has to be done in the variables taken by the system.

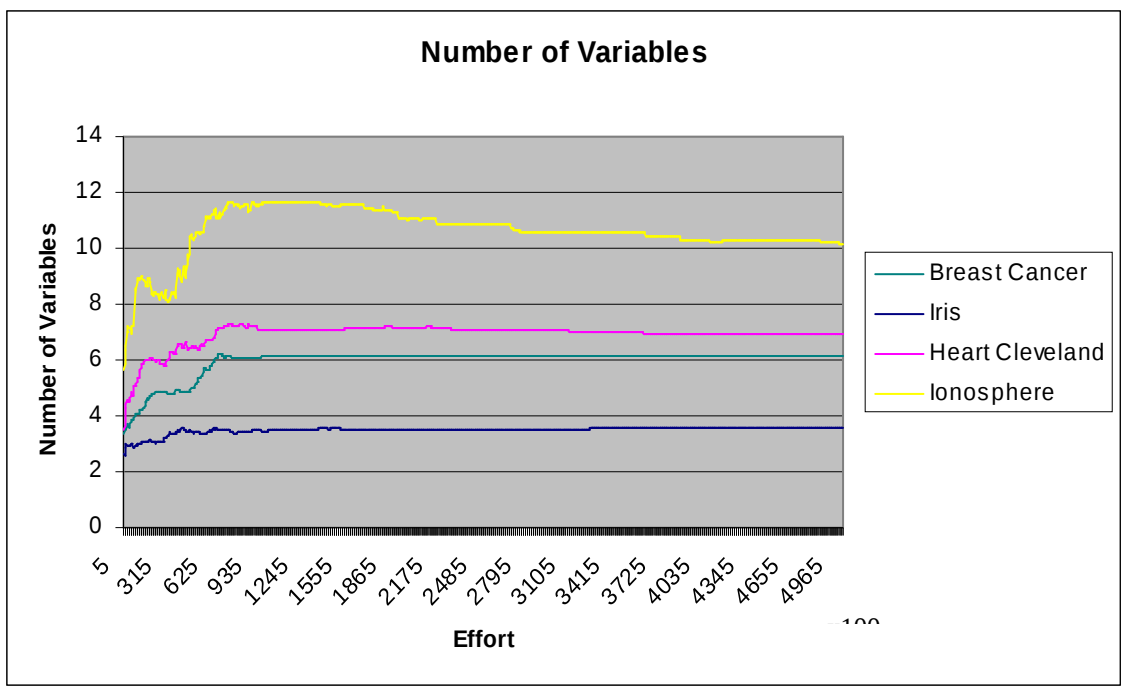

**Fig 4 Average number of used features**

This reduction of the number of input features is graphically shown on Fig. 4. This figure shows an average of the number of features used by the different networks in the execution process until a maximum of 500.000 pattern file evaluations.

*6.4. Optimization of the networks*

Another important feature of this technique is that it makes an optimization of the networks. As was already explained, this optimization depends on the parameter that penalizes the networks on the number of neurons they have. This forces the system develop networks with a low number of neurons. An example of this feature can be seen on figure 2, where a network that solves the ionosphere problem has been found. This network is much simpler than the classical multilayer perceptrons (MLP). Table 17 shows a comparison between the network found here and the ones that could be used with the classical system, i.e., with a total connectivity between the neurons of one layer and the following (see figure 3). As can be seen on this table, the network found in this work presents an important improvement from the point of view of the number of neurons and connections.

Table 17. Comparison between the network found with this method and the classical multilayer perceptrons

| Method | Number of hidden neurons | Number of connections |
|---|---|---|
| MLP | 25 | 875 |
|  | 20 | 700 |
|  | 15 | 525 |
|  | 3 | 105 |
| Proposed here | 3 | 13 |

Another example can be seen with the iris flower problem. Previous works have been done with this problem in order to analyze the resulting networks . Table 18 shows a summary of the accuracies presented in that work, along with the number of networks and connections, in comparison with the ones obtained with the classical method .

As can be seen on tables 17 and 18, the networks found by this system are much simpler than the ones used with the classical method.

Table 18. Comparison between the network found with this method and the ones found with the classical method in the iris flower problem

| Method | Hits | Accuracy | Hidden neurons | Connections |
|---|---|---|---|---|
| Rabuñal 1999 | 148 | 98.66 % | 5 | 35 |
| Rivero 2007 | 149 | 99.33 % | 3 | 15 |
|  | 148 | 98.66 % | 1 | 11 |
|  | 147 | 98 % | 1 | 9 |
|  | 146 | 97.33 % | 1 | 10 |
|  | 145 | 96.66 % | 1 | 10 |

*6.5. Architectures*

An additional advantage of this system is that the layers of the networks do not have the limitations of the networks built by the traditional networks, but internally the network can have any type of connectivity. In the particular case of this work, this connectivity has been limited only to obtaining networks without recurrent connections, as was previously explained.

This ability of having any type connectivity is a great advantage with the ANNs that possess a traditional architecture, because they present many difficulties when being analysed. The networks developed by this system, however, as they have underwent an optimization process, have a low number of neurons. This makes them much easier to analyse to discover, for example, which variables participate in a determined output of the network.

For example, table 17 shows a summary of the architectures found of the networks that solved the iris flower problem with different accuracies in comparison with a classical network used in previous works. This network is very difficult to analyse due to the great amount of neurons and connections. However, the networks described on table 16 are much simpler and can be more easily analysed. In fact, they were analysed and the conclusions at the end of the section 8.3 about the inputs of the problem were obtained.

**7. CONCLUSIONS**

In this paper a technique is presented with which it is possible to generate ANNs by means of a hybrid system that combines two Evolutionary Computation techniques: Genetic Programming and Genetic Algorithm. This technique was designed for use with biomedical data, and the results show its great performance in different problems, which will be very useful in biomedical environments.

Also, the results obtained in other databases show that this technique can also be used in other data.

Section 8 presents a set of features that the system described here has, and can be summarized as following:

- The results obtained make this tool, on average, better than the rest of the tools used for the comparison.
- This system is completely independent from the expert. The expert does not have to do any effort to execute it, even for tuning the system parameters. The finding of the optimal architecture and weight values is completely automated.
- The system performs a discrimination of the input variables, which allows the obtaining of knowledge about the problem domain.
- The networks obtained by the system have been optimized so they contain a minimal set of hidden neurons.
- The architectures found by the system can have any type of connectivity. This allows a better and much easier analysis of the networks, which a classical architecture does not allow.

Therefore, the technique presented in this paper is a powerful tool that can develop simple ANNs without human intervention.

## 8. FUTURE WORKS

One interesting research line is the study of the adaptation of this system so it can be used by a GP algorithm based on graphs. In this way, it would be possible to avoid the use of a list and special operators for referencing an already existent neuron.

The parameters of the evolutionary algorithm can also be further researched. Recent works point that good results could be obtained when not using crossover, i.e., using only mutation .

Another interesting aspect would be to research the utilization of GP distributed systems distributed for generating ANNs, and in this way, improve the performance of the system with respect to the computational load necessary for obtaining good results.

## 9. ACKNOLEDMENT

The development of the particular experiments in this paper were made with thanks to the support of the "Centro de Supercomputación de Galicia (CESGA)".

The Cleveland heart disease database was available thanks to Robert Detrano, M.D., Ph.D., V.A. Medical Center, Long Beach and Cleveland Clinic Foundation.